\pdfoutput=1

\documentclass[11pt]{article}

\usepackage[final]{faaf24}
\usepackage{mdframed}
\usepackage{times}
\usepackage{latexsym}
\usepackage{verbatim}
\usepackage{amsmath}
\usepackage{longtable}
\usepackage{color,soul}
\usepackage{xltabular}
\usepackage{CJKutf8}
\usepackage{multirow}
\usepackage{booktabs}

\usepackage[utf8]{inputenc}

\usepackage{microtype}
\usepackage{graphicx}
\usepackage{inconsolata}

\title{FaaF: Facts as a Function for the evaluation of generated text}

\author{
    Vasileios Katranidis \hspace{1.5cm}\quad Gabor Barany \hspace{1.5cm}  \\
    IMMO Capital \\
    \texttt{\{vasileios.katranidis, gabor.barany\}@immo.capital}
}

\begin{document}
\begin{CJK*}{UTF8}{gbsn}
\maketitle

\begin{abstract}
  The demand for accurate and efficient verification of information in texts generated by large language models (LMs) is at an all-time high, but remains unresolved. Recent efforts have focused on extracting and verifying atomic facts from generated texts via prompting LM evaluators. However, we demonstrate that this method of prompting is unreliable when faced with incomplete or inaccurate reference information. We introduce Facts as a Function (\textbf{FaaF}), a new approach to the fact verification task that leverages the function-calling capabilities of LMs. \textbf{FaaF} significantly enhances the ability of LMs to identify unsupported facts in texts, while also improving efficiency and significantly lowering costs compared to prompt-based methods. Additionally, we propose a framework for evaluating factual recall in Retrieval Augmented Generation (RAG) systems, which we employ to compare prompt-based and \textbf{FaaF} methods using various LMs under challenging conditions.

\end{abstract}
\section{Introduction}
The adoption and transformative impact of large language models (LMs) across industries are significantly driven by their application in knowledge-intensive tasks.
\begin{figure}
\vspace*{-1 cm}
    \centering
    \includegraphics[width=1\linewidth]{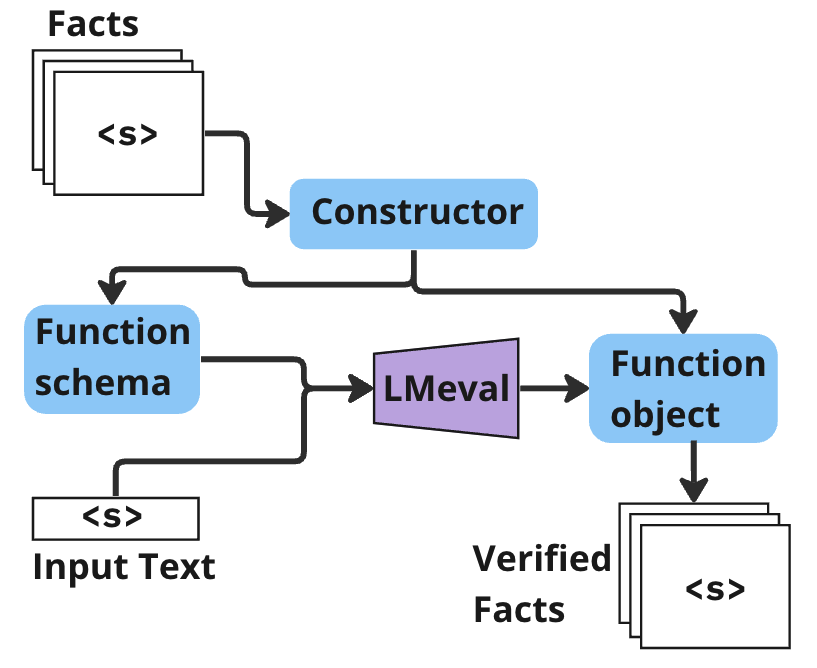}
    \caption{An overview of FaaF, a constructor dynamically creates a function object from a set of fact statements. Function calling allows LMeval to verify all facts within a single call when provided with an input reference text. FaaF significantly reduces the error rate in identifying unsupported facts compared to prompting whilst reducing the number of LMeval calls and output tokens by more than 5 times.
}
\end{figure}
In these applications, Retrieval Augmented Generation (RAG) is often used to integrate out-of-training knowledge to the LM \citep{lewis2021retrievalaugmentedgenerationknowledgeintensivenlp}.

Considering the importance of factual accuracy of generated text in this setting, a significant body of work has focused on automated ways for obtaining it. At a high level, two distinct schools of thought emerge - shaped by the intended use case, the available test datasets and required scale. First, authors who address \textit{factual precision} —the truthfulness of each statement in a generated text— in both RAG and non-augmented LM generation scenarios \citep{chen2022generating, zhang2023interpretable,gao2023rarr, lee2023factuality, min2023factscore, azaria2023internal, yuan2021bartscore, fu2023gptscore, wang2023chatgpt, kadavath2022language, manakul2023selfcheckgpt}. Since a generated text may include both accurate and inaccurate statements, a common approach is to initially extract a number of fact statements from the generated text and verify them individually by prompting an LM evaluator model with appropriate references. Secondly, other studies \citep{cuconasu2024power,liu2023lost,kandpal2023large,mallen2023trust} attempt to evaluate LMs and RAG systems in terms of exact matching any one of a set of predefined, correct answers in the generated text.

However, as recognised by \citet{cuconasu2024power}, exact matching of ground truth answers in the generated text is prone to false negatives since the ground truth information might be present but phrased differently. This issue is exacerbated  when the grounding information is longer than a few words making the conditional probability of an exact match very low. Secondly, merely the presence of specific words or phrases in the generated text is not sufficient to evaluate the truthfulness of more sophisticated statements.

Although the exact match method can be viewed as binary form of factual recall, where the existence of a single accepted answer in the generated text signals success, it faces serious limitations in the verification task. Considering the importance of \textit{factual recall} —the extent to which all the information required to sufficiently answer a question is included in the generated text— in the RAG scenario, there is an apparent lack of work on its practical measurement. Factual recall directly captures the performance of retrieval and generation of a RAG simultaneously and closely reflects the central purpose of the system. In this use case, recall is more important than precision since a generated response can be factually precise but given the wrong context may be irrelevant to the posed question. 

Further, current approaches relying on verification of each fact independently can be prohibitively costly in time and resources. Specifically, RAG systems include many moving parts (knowledge base, retrieval, prompt formulation, LM) and require substantial tuning \citep{es2023ragas} therefore the efficiency and speed of the evaluation task is a requirement for practical usage.

To address the gaps above, we  make the following contributions: 
\begin{enumerate}
    \item We introduce Facts as a Function (FaaF), a new fact verification formulation which results in significantly more accurate and efficient verification compared to current approaches relying on prompting.
    
    \item We probe into the performance of fact verification formulations in conditions of highly incomplete or inaccurate generated text in a controlled manner. To achieve that, we augment WikiEval\footnote{https://huggingface.co/datasets/explodinggradients/WikiEval}\citep{es2023ragas} which features question/answer pairs with answers of variable factual quality which enable simulating deficient RAG responses. We find that prompt-based fact verification faces serious challenges in identifying unsupported facts in the presence of inaccurate or incomplete generated text.

    \item We open source FaaF\footnote{https://github.com/vasiliskatr/faaf}, the factual recall evaluation framework and the augmented WikiEval dataset (WikiEvalFacts)\footnote{https://huggingface.co/datasets/Vaskatr/WikiEvalFacts} to help the community include factual recall in the RAG optimisation and ultimately build more reliable systems.
\end{enumerate}

\section{Related Work}
Recently, \citet{min2023factscore} used prompt variations and an aggregate non-parametric probability of the tokens in a fact statement to directly verify individual facts extracted from LM generated biographies. They compare their method with human evaluation and find low error rates when retrieving the ground truth for the evaluated fact.

\citet{zhang2023interpretable} propose self-measuring factuality by the LM via a few-shot prompting method combined by generated facts pertinent to the statement in question. They argue that while leveraging facts from a knowledge base is more dependable, its effectiveness is confined to the scope of the knowledge base and the quality of retrieval. Conversely, self-evaluating with generated facts offers more flexibility but risks introducing inaccuracies.

\citet{li2023halueval} indicate that LMs have difficulty identifying non-factual information with standard prompting strategies and report improvement using Chain of Thought (CoT).

\citet{azaria2023internal} also find that fact verification by prompting is insufficient and propose to train a classifier on the hidden layer activations of open source LMs to predict the truthfulness of a generated statements. However, current leading commercial models are lacking layer activation access and so require alternative methods.

Another approach in the same spirit is to look at the probabilities of each generated token as an indicator of LM confidence and truthfulness of the generated text with the view that low LM confidence is a proxy for incorrect statements \citep{yuan2021bartscore}.

\citet{fu2023gptscore} build on the concept of utilising token probabilities, introducing a self-evaluation framework for LMs. This framework leverages few-shot prompting to evaluate various instructed aspects of LM responses, such as factuality, fluency, interest, among others.

\citet{manakul2023selfcheckgpt} propose SelfCheckGPT which automates the detection of factual errors in LM outputs through statistical analysis of multiple responses to the same prompt, without external knowledge sources. This is again, an expression of the general idea that the probability distribution of the generated response is indicative to the confidence on its truthfulness. So similar to  \citet{yuan2021bartscore},  SelfCheckGPT makes this assessment post LM-generation by sampling multiple answers on the same prompt thereby removing the requirement of access to the token probabilities or layer weights and making this approach applicable to closed models.

\citet{aly2021feverous} use a Roberta encoder with a linear layer to learn and predict the fact label given text evidence.

\citet{wang2023chatgpt} describe a method where the LM is prompted directly to score an answer's specific aspect from 0 to 100 or rates it on a 5-star scale, yielding notable results. However, this approach's effectiveness heavily depends on the prompt's design.

\citet{zhang2020bertscore} attempt a flexible self-evaluation of  generated text using reference answers (BertScore). BertScore calculates a similarity score between tokens in the generated and reference sentences using contextual embeddings. The key benefit being that there is no reliance on exact matching between generated and reference text. Nevertheless, a high semantic score at sentences level does not guarantee factual precision, especially when the information examined is not contextual and depends only on a small number of tokens (E.g. date).

The work of \citet{zhao2019moverscore} also relies on contextual embeddings but their approach allows for an intentional bias towards precision or recall via reformulating the semantic similarity between generated and reference text as an optimisation problem of finding the minimum effort to transform between the two.

\citet{kadavath2022language} observe that LLMs offer well-calibrated probabilities for self-evaluation via constraining the LM response into multiple-choice and True/False questions. This work highlights that simply requiring discrete response options prior to text generation can aid the response calibration by effectively narrowing the available distribution of next tokens --- which would alternatively include many semantically overlapping paraphrases.

Lastly, recent work on the factual accuracy of LMs and RAG systems \citep{cuconasu2024power,liu2023lost,kandpal2023large,mallen2023trust} took the approach of using the NaturalQuestions-Open (NQ-open)\footnote{https://ai.google.com/research/NaturalQuestions} dataset \citep{kwiatkowski-etal-2019-natural} and calculate accuracy by judging whether any of the ground truth answers (NaturalQuestions annotations) appear in the generated text via exact matching. NQ-open is a large scale dataset which comprises historical Google search queries and their human-annotated answers sourced from Wikipedia. Even though NQ-open is valuable for its extensive scope and domain-agnostic nature, fact-verification via exact matching faces serious challenges \citep{cuconasu2024power} and a more advanced verification of answers is left for future research.

\section{Facts as a Function}
Facts as a Function (FaaF) is a streamlined fact verification method using function calling for multi-fact assessment. 

Assuming a set of fact statements to be verified,  we construct a facts-specific function object and a parsing function. Since the created function object contains all the input facts as arguments, we perform verification to the set of facts as a unit. An example of the JSON representation of a function object containing the first fact can be seen bellow:\\
\begin{verbatim}
{'properties':
    ‘fact_0:{
    'description':"It
    is clear from the passage
    that Pope Benedict XVI became
    the head of the Catholic Church
    and sovereign of the Vatican
    City State on April 19, 2005.
    Respond by using one of the accepted
    Enum types.",
   	'enum': ['True', 'False'],
   	'type': ‘string’
    },
…

 },
'required':
	[‘fact_0',
 	‘fact_1’
 	 …,
 	‘fact_n’]
'title': 'FactChecker',
'type': ‘object'}
\end{verbatim}

\noindent\textbf{Code over natural language}\\
We propose that by using the function calling ability of the LM, we enforce a more formal mode of token generation compared to natural language. Function calling can be viewed as prompting the LM to generate code (the function arguments JSON) but this has important effect in the generation. First, due to the strict nature of code syntax compared to natural language, gradients during training are expected to be steeper which ultimately leads to better model calibration when generating code (e.g. arguments for a function). This manifests as better adhering to the expected output and responding with lower stochasticity.

By leveraging the metadata of function arguments, type annotations and tailored instructions we can constrain the LM to the accepted modes of response more effectively than prompting. Ultimately, we can avoid relying on exact matching to interpret the LM's response---which can prove detrimental as we demonstrate in the results of this paper. 

Additionally, type annotations can be combined with custom types to essentially convert a function argument into a classification result to a multiple-choice question. Building on the findings of \citet{kadavath2022language}, who established that LMs show well calibrated probabilities when presented with multiple choice questions, we propose that using the function argument's type annotations to convey the accepted LM responses is a step further in the same direction.\\

\noindent\textbf{Generated text as a unit}\\
A function definition can encapsulate all the facts statements which need to be verified for a piece of long-form text. Therefore, we move away from the concept that each fact should be verified individually via a fact-specific prompt and we propose instantiating a function per LM-generated text which needs to be factually assessed. This results in a reduction of cost and time for fact-verification which is proportional to the number of facts which would otherwise needed to be assessed individually, as seen in \citet{chen2022generating,gao2023rarr, min2023factscore,lee2023factuality}.\\

\noindent\textbf{Outsourcing judgement from the LM to the function}\\
Using function objects to communicate with the LM, enables access to a multitude of tools and further processing we can execute on LM’s output. This strategy permits us to delegate certain deterministic judgments away from the language model. We demonstrate this capability by mapping a range of LM responses into a binary format (\texttt{True} / \texttt{False}). In doing so, the calibration of the LM response is enhanced as we provide a more accurate representation of the spectrum of potential outcomes than a simple True/False dichotomy. The underlying intuition is that ultimately, we can afford to ask simpler and clearer questions to the LM which can be answered more reliably and further process the LM output into a final response.\\

\noindent\textbf{Definition}\\
We aim to present the facts to the LM as a callable function. Let \(S\) be the list of fact statements as strings to verify.
\begin{equation*}
    S=[s_{1}, s_{2}, \ldots, s_{n}]
\end{equation*}

A constructor function \(C\) then maps the input list of facts \(S\) and control parameters \(P\) to an function object \(O\) with arguments \(f\).
\begin{equation*}
C(S,P) \to  O\left(f_1, f_2, \ldots, f_n\right)
\end{equation*}

Each argument from $(f_1, f_2, \ldots, f_n)$ corresponds to a fact statement and is further parameterised by $P$. Control parameters $P$ include the methods, argument properties and metadata which are injected into resulting object $O$. Such methods can describe for example a desired post-processing step on the values in the arguments $(f_1, f_2, \ldots, f_n)$. Before passing object $O$ to a language model, we convert it to a JSON or XML representation---depending on the LM's function calling requirements:
\begin{equation*}
    J(O) \to \text{JSON}_{O}
\end{equation*}

Let $M$ be the language model used for fact verification (LMeval). The input of  $M$ is a concatenation of $JSON_{O}$, a prompt  $q$ which instructs  $M$ to utilise  $O$ and the input text  $x$ which is to be assessed for factuality with respect to the given facts  $S$.
\begin{equation*}
    M(\text{JSON}_O, q, x) \to o_x
\end{equation*}

$M$ responds with the output $o_{x}$ which is passed to a parsing function $G_{M}$ which adheres to particular response schema of $M$ and invokes $O$ by assigning values on its arguments, yielding $\acute{O}$. The values being assigned to each of $(f_1, f_2, \ldots, f_n)$ being the verification result of the underlying fact statement. The generated function arguments undergo type validation and modification, if required, to ensure that the schema and type annotations are respected.
\begin{equation*}
    G_{M}(o_{x})\to \acute{O}
\end{equation*}
Finally, the fact-verification process can be expressed as
\begin{equation*}
\acute{O}=G_{M}(M(\text{JSON}_O,q,x)).
\end{equation*}

\begin{figure*}
    \centering
    \includegraphics[width=\textwidth]{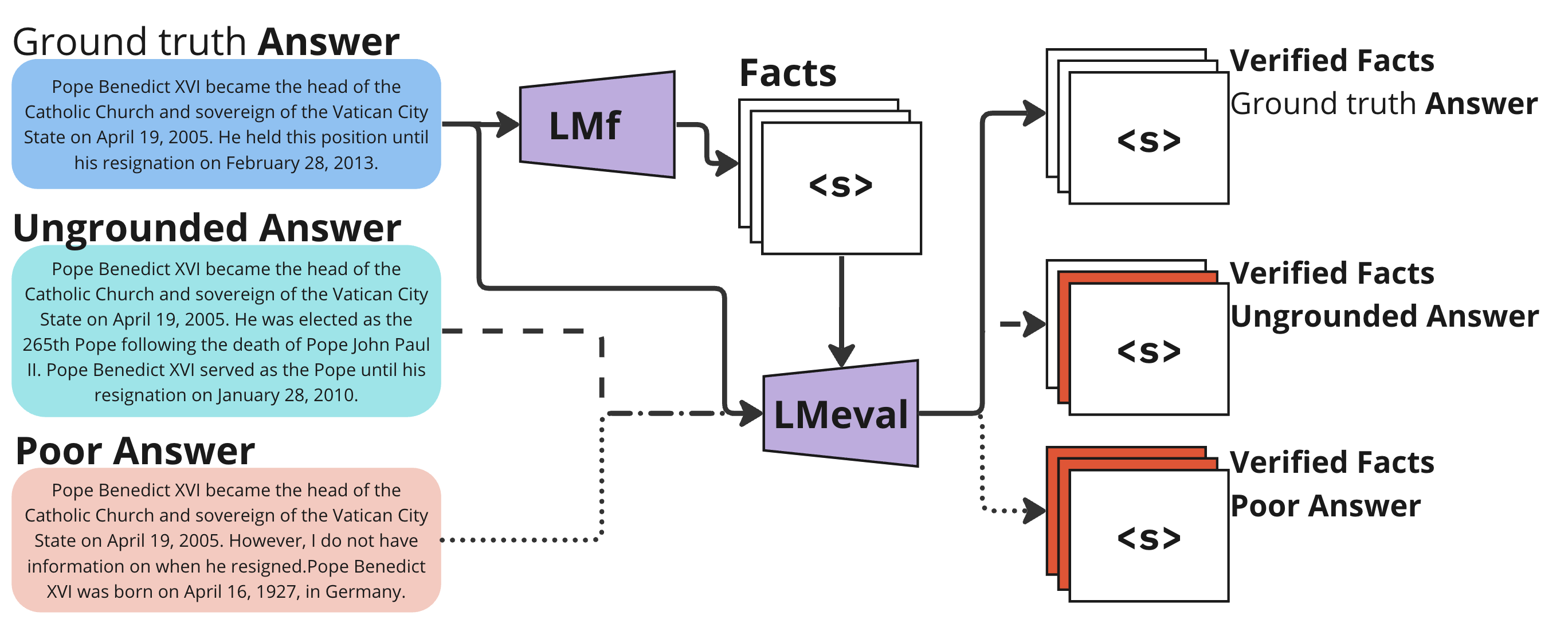}
    \caption{Overview of the factual recall evaluation for RAG. Given a set of ground truth \textbf{Answers}, facts are extracted via LMf. The hypothesized responses of the RAG (in this instance \textbf{Ungrounded Answer} and \textbf{Poor Answer}) are then tested for recall against the extracted facts.}
\end{figure*}

\section{Assessment of fact verification formulations in the RAG setting}
Figure 2 outlines the factual recall evaluation framework which also serves as the experimental setup which we use to compare fact verification formulations with each other. Starting from the ground truth \textbf{answer} containing the desired information to fully address the posed \textbf{question}, we derive a set of fact statements using a fact-generator LM (LMf). We then use these derived facts to evaluate each of the other answer variants in WikiEval for their factual recall via LMeval. In this manner, each answer is evaluated with respect to the information that is expected from it. It is easy to see how this framework could be applied in the RAG setting where different configurations or model choices impact the quality of the final response.\\

\noindent\textbf{Dataset}

\noindent In order to probe into performance of automatic fact verification methods we chose to work with the WikiEval dataset \citep{es2023ragas} which features three versions of answers, of variable quality, to questions on Wikipedia articles. Specifically, for each question, there is an \textbf{answer} (referred to as ground truth \textbf{answer} in this paper for clarity), \textbf{ungrounded answer }and \textbf{poor answer}. All answers have been generated with \texttt{GPT-3.5-turbo-16k}.

The ground truth \textbf{answer} is generated by providing the LM with the correct context from the respective Wikipedia page. The \textbf{ungrounded answer} is generated by not providing any context to the LM. Finally, the \textbf{poor answer} is generated by instructing the LM to give an incomplete answer to the given question. An example of each answer type can be seen in Figure 2. Generally, \textbf{ungrounded answers} contain false, incomplete and redundant information with respect to the ground truth \textbf{answers}. On the other hand, \textbf{poor answers} contain primarily incomplete information or lack information altogether, compared to ground truth \textbf{answers} i.e. no evidence for support or rejection for ground truth facts.

This dataset enables us to assess the impact of quality and completeness of different answer variants to the performance of fact-verification. In that way, we test closely the ability of different fact verification methods and LMs to identify unsupported facts when presented with (i) incorrect, (ii) indirectly relevant and (ii) incomplete information with some degree of distinction.\\

\noindent\textbf{Fact generation}\\

\noindent To prepare the WikiEval dataset, we initially generate fact statements that fully capture the information from the ground truth \textbf{answers}, followed by manual annotation of the generated facts for each answer variant. We call the resulting augmented dataset WikiEvalFacts. We use \texttt{gpt-4-turbo} to generate facts from the questions and ground truth \textbf{answers} from the WikiEval dataset (QA pair) using the following prompt:
\begin{quote}
    \textit{Convert the given passage into a list of short facts which specifically answer the given question. }

\textit{Make sure that the facts can be found in the given passage.}

\textit{The facts should be coherent and succinct sentences with clear and simple syntax.}

\textit{Do not use pronouns as the subject or object in the syntax of each fact.}

\textit{The facts should be independent to each other.}

\textit{Do not create facts from the passage which are not answering the given question.}

\textit{Add a "-" before each fact.}

\textit{Passage: [ground truth answer]}

\textit{Question: [question]}
\end{quote}

 Fact generation via the prompt above results in a variable number of facts for each ground truth QA pair which depends on the length and information density in the processed ground truth \textbf{answer}. This process yielded 281 individual facts, which were annotated for each answer type (thus 843 annotated facts in total considering ground truth \textbf{answer}, \textbf{ungrounded answer} and \textbf{poor answer}) with an average of 5.6 fact statements generated for every QA pair. The prompt has been designed to ensure that the generated facts are complete sentences, understandable independently of each other or any external references.\\
 \begin{table}[h!]
  \centering
  \begin{tabular}{lccc}
    \toprule
      &\textbf{Factual Accuracy} \\
     \textbf{Answer Type} & Human Evaluation\\
    \midrule
    \ Ground Truth Answer & 100 \\
    \ Ungrounded Answer & 30.6\\
    \ Poor Answer & 8.5 \\
    \bottomrule
  \end{tabular}
  \caption{Factual accuracy of the facts derived from the ground truth \textbf{Answer} of WikiEval from human evaluation.}
\end{table}

\noindent\textbf{Human fact-verification}

\noindent We outsource the fact verification of the generated facts against the triplet of answers in WikiEval (ground truth \textbf{answer}, \textbf{ungrounded answer} and \textbf{poor answer}) to human evaluators. In this manner we build a ground truth evaluation for each answer type, enabling us to assess the effectiveness of automated fact-verification methods against it. The accuracy from the human fact verification can be seen in Table 1 where the factual accuracy of ground truth \textbf{answers} is 100\% since all the generated facts are True by design. The deterioration of the \textbf{ungrounded answer} and \textbf{poor answer} relative to the ground truth \textbf{answer} is evident.\\

\noindent\textbf{Prompt fact-verification}\\
\noindent Following \citet{min2023factscore}, we use a prompt and the respective answer variant as context to verify a single fact at a time with LMeval:

\begin{quote}
    \textit{Passage: [answer]\\\\
Considering the given passage, the claim 	[fact] is True or False?}
\end{quote}

\noindent\textbf{Facts as a function}

\noindent For each set of fact statements which encapsulate a ground truth \textbf{answer}, we construct a facts-specific function object as seen in section 3. Each function argument includes metadata which can be used to pass instructions, type annotations and the fact statement to be verified itself. In addition to the function object, we pass the following prompt:

\begin{quote}
    \textit{Consider the given passage and assign the correct values in the fact checker function.\\\\
    Passage: [answer]}
\end{quote}

The \textit{answer} in the prompt is the input text we want to evaluate against the respective facts which have been previously derived by the ground truth \textbf{answer} in our dataset. After LMeval generates a response, the parsing function is used to invoke the function object by supplying the arguments parsed from the LMeval's response.\\

\noindent We test the following configurations:\\

\textbf{FaaF(T/F)}\\
A function object with arguments which only accept \texttt{True} or \texttt{False} as a response from LMeval. As seen from the JSON example above, these are specified as custom type annotations (enum).\\

\textbf{FaaF(T/F/N)}\\
A function object with arguments which only accept \texttt{True}, \texttt{False} or \texttt{Not clear from the given passage} as a response from LMeval. In this scenario, further processing inside the function object will map \texttt{Not clear from the given passage} to \texttt{False} after invocation. This is an example of applying a simple processing step on the LM output, post-generation. The intuition behind this configuration is that the rejection of a claim based on contradicting evidence is conceptually different to the rejection of a claim based on absence of evidence and we help the LMeval's calibration by providing a clear response option for each.\\

\noindent\textbf{FaaF(T/F)+citation}\\
In this instance we construct a function object with two arguments for each input fact. One argument for the factual evaluation and one argument where we instruct the LMeval to generate an exact excerpt from the input text which directly supports the fact in question (i.e. citation). We place the citation argument prior to the factual evaluation argument so that LMeval is made to first try and find a supporting citation from the input text before verifying the fact that is being assessed. Similarly, the intuition here is that by asking LMeval to search and retrieve a specific citation from the input text which supports a specific fact, it will result in a better calibrated verification of the respective fact statement.\\

\noindent\textbf{FaaF(T/F/N)+citation}\\
In this configuration we combine the two approaches outlined above to explore their combined effect. In detail, we construct the function object to include citation arguments and we define \texttt{True}, \texttt{False} or \texttt{Not clear from the given passage} as accepted responses.\\\\
\textbf{Language models (LMeval)}\\
We use established commercially available models which support function calling \texttt{gpt-4-turbo}, \texttt{gpt-3.5-turbo}, \texttt{claude-3-opus} and \texttt{claude-3-sonnet}. We also examined \texttt{mistral-Large} but it was excluded from the results in this paper due to its high failure rate of over 80\% in some cases in generating appropriately formatted responses, rendering its results non-contributory to the discussion. It is important to note that we did not allow models to retry in case of a failed invocation of the function object due to formatting. FaaF introduces strict constraints on the expected LM response and by permitting only one attempt, we also assess the LM's proficiency in formatting their response, as well as verifying factual accuracy. 

We kept the system prompts for the GPT models unchanged but modified Claude LMs system prompts to incorporate a 1-shot example of simple function calling. This adjustment is a result of following the official function-calling recommendations of the two model families at the time of writing this paper.\\

\noindent\textbf{Metrics}\\
The capacity of the LM to return a correctly formatted response for function calling is distinct to their ability for accurate fact verification. The verification accuracy metrics are calculated considering only the correctly formatted LM responses (first attempt only). In doing so, we ensure that the comparison of the LMs' fact verification ability is not influenced by their capacity to format responses correctly --- which is discussed separately.
\indent We use \textbf{Error Rate} (ER) between the human fact-verification and the fact verification formulation as the main indicator of verification accuracy.
\indent We also use \textbf{F1micro} score (F1m) as defined in \citet{min2023factscore} to measure the successful identification of unsupported facts and probe further into the individual fact verification. It should be noted that F1 scores explicitly depend on the class ratio (T/F) via precision and recall. For that reason F1m scores should be compared across fact verification approaches in the same answer category (where the T/F ratio is preserved) in Table 2, and not across answer categories.

\begin{table*}[ht]
\centering
\setlength{\tabcolsep}{3pt}
\begin{tabular}{@{}llccccccccc@{}}
\toprule
&  & \multicolumn{3}{c}{(ground truth)\textbf{ Answer}} & \multicolumn{3}{c}{\textbf{Ungrounded Answer}} & \multicolumn{3}{c}{\textbf{Poor Answer}} \\
\cmidrule(lr){3-5} \cmidrule(lr){6-8} \cmidrule(l){9-11}
& \textbf{Facts Formulation}  & \textbf{N/A} & \textbf{ER} & \textbf{F1m} &  \textbf{N/A} & \textbf{ER} & \textbf{F1m} & \textbf{N/A} & \textbf{ER} & \textbf{F1m}   \\

\midrule

\multirow{5}{*}{\rotatebox[origin=c]{90}{gpt-3.5-turbo}} & Prompt(T/F) & 0/281 & 1.4 & 0$^{*}$ & 0/281 & 27.4 & 76.5 & 16/281 & 55.3 & 56.9 \\

& FaaF(T/F) & 0/281 & 1 & 0$^{*}$ & 0/281 & 23.1 & 81 & 0/281 & 17.4 & 89.6 \\

& FaaF(T/F/N) & 0/281 & 1 & 0$^{*}$ & 0/281 & 25.9 & 78 & 0/281 & 18.5 & 88.8 \\

& FaaF(T/F)+citation & 0/281 & 1 & 0$^{*}$ & 26/281 & 28.2 & 75.3 & 71/281 & 15.7 & 90 \\

& FaaF(T/F/N)+citation & 0/281 & 1.4 & 0$^{*}$ & 26/281 & 31.3 & 71 & 31/281 & 23.2 & 85.5 \\

\midrule

\multirow{5}{*}{\rotatebox[origin=c]{90}{claude-3-sonnet}} & Prompt(T/F) & 0/281 & 1 & 0$^{*}$ & 0/281 & 42.7 & 58 & 2/281 & 77.7 & 26.4 \\

& FaaF(T/F) & 0/281 & 1.7 & 0$^{*}$ & 0/281 & 28.8 & 75.9 & 0/281 & 14.9 & 91.1 \\

& FaaF(T/F/N) & 0/281 & 2 & 0$^{*}$ & 0/281 & 27.7 & 76.9 & 0/281 & 14.2 & 91.6 \\

& FaaF(T/F)+citation & 0/281 & 0.7 & 0$^{*}$ & 0/281 & 30.6 & 73.9 & 0/281 & 14.5 & 91.4 \\

& FaaF(T/F/N)+citation & 0/281 & 1 & 0$^{*}$ & 0/281 & 27.4 & 76.7 & 0/281 & 14.9 & 91.3 \\

\midrule

\multirow{5}{*}{\rotatebox[origin=c]{90}{gpt-4-turbo}} & Prompt(T/F) & 0/281 & 1.7 & 0$^{*}$ & 1/281 & 24.2 & 80.1 & 9/281 & 44.1 & 68.4 \\

& FaaF(T/F) & 0/281 & 1 & 0$^{*}$ & 0/281 & 22.4 & 82 & 0/281 & 8.5 & 95.1\\

& FaaF(T/F/N) & 0/281 & 1.4 & 0$^{*}$ & 0/281 & 17.7 & 86.3 & 0/281 & 6.7 & 96.2\\

& FaaF(T/F)+citation & 0/281 & 0.7 & 0$^{*}$ & 6/281 & 15.6 & 88.2 & 15/281 & 7.5 & 95.8\\

& FaaF(T/F/N)+citation & 0/281 & 1 & 0$^{*}$ & 6/281 & 16 & 87.7 & 0/281 & 9.2 & 94.8\\

\midrule

\multirow{5}{*}{\rotatebox[origin=c]{90}{claude-3-opus}} & Prompt(T/F) & 0/281 & 1.7 & 0$^{*}$ & 0/281 & 42.3 & 58.8 & 2/281 & 76.3 & 28.7 \\

& FaaF(T/F) & 0/281 & 3.2 & 0$^{*}$ & 0/281 & 15.3 & 88.8 & 0/281 & 6.7 & 96.2\\

& FaaF(T/F/N) & 0/281 & 3.5 & 0$^{*}$ & 0/281 & \textbf{14.5} & \textbf{89.3} & 0/281 & \textbf{4.9} & \textbf{97.2}\\

& FaaF(T/F)+citation & 0/281 & \textbf{0.3} & 0$^{*}$ & 0/281 & 24.1 & 80.8 & 0/281 & 5 & 97\\

& FaaF(T/F/N)+citation & 0/281 & 0.7 & 0$^{*}$ & 0/281 & 20.9 & 83.6 & 0/281 & 7.8 & 95.6\\

\bottomrule
\end{tabular}
\caption{Results on Error Rate (ER) along with F1micro estimated by each fact formulation. F1micro measures accuracy and recall in predicting the false facts. Bold case indicates best performance. $^{*}$F1micro is 0 in the Ground Truth Answer cases because there are no False facts in the ground truth evaluation in this category. T/F indicates True/False and T/F/N are True/False/Not clear from the given passage respectively. N/A indicates the Not answered i.e. the number of facts that the LM did not respond on their validity due to errors in its response format.}
\end{table*}

\section{Results}
Table 2 presents the non-answer rate (N/A), Error Rate and F1m from the examined fact verification formulations and LMs, across the answer categories which are examined.\\

\noindent\textbf{Prompting for fact verification is not reliable in cases of incorrect or incomplete information}. Although in the case of ground truth \textbf{answers} \textbf{Prompt(T/F)} ER is in the low percentage points in all LMs—in line with what is reported in \citet{min2023factscore} when retrieval is enabled—we see a sharp rise when we try to verify facts in text from the \textbf{ungrounded answers} and a further deterioration with \textbf{poor answers} with ER exceeding 50\% and 70\% with GPT and Claude LMs respectively. Although the failure mechanisms are distinct between Claude and GPT, the performance in both cases indicates that prompting coupled with word-matching is not a suitable approach for fact verification in text with unknown information quality and completeness.

In the case of GPT LMs, the high error rates are attributed to an overall overestimation of the factual truthfulness of a fact statement given a reference text which does not support it. Regarding Claude LMs, the exceedingly high ER is primarily due to the erroneous parsing of the LM's response. In detail, on many occasions the verification response from Claude LMs contains both \texttt{True} and \texttt{False} words, which exposes the fragility of word-matching as a means of parsing the LM's response. For example, Claude responses frequently include phrases like \textit{"To determine if the claim is true or false based on the given passage..."} which mislead the interpretation of the response when using word matching.\\

\noindent\textbf{FaaF outperforms prompting in fact verification accuracy}. Formulations which leverage function calling demonstrate a notable improvement in ER and F1micro particularly in the cases of \textbf{ungrounded} and \textbf{poor answer} with all LMs. The improvement is most pronounced in the \textbf{poor answer} category where all models show lower ER scores from 30 to 70 percentage points compared to prompting, denoting a paradigm shift in the model's ability to identify unsupported facts in the examined text. In addition, the larger LMs (\texttt{claude-3-opus} and \texttt{gpt-4-turbo}) preform notably better in the \textbf{poor answer} category than their respective inferior versions (\texttt{claude-3-sonnet} and \texttt{gpt-3.5-turbo}).\\

\noindent\textbf{Rejecting facts is more difficult in the presence of tangentially relevant information when using FaaF}. It is evident from Table 2 that FaaF approaches struggle more with \textbf{ungrounded answers} than with \textbf{poor answers}. The key difference between the two answer types is that \textbf{ungrounded answers} typically contain incorrect or tangentially relevant information to the facts that are examined whilst \textbf{poor answers} usually lack information altogether. This is because \textbf{ungrounded answers} are generated by an LM without any context thus, depending on the popularity and prevalence of the theme of the question, a degree of factual hallucination can be expected. It is apparent that, from the perspective of fact verification, presence of indirectly relevant (but not useful) information in the generated text is more dangerous than lack of information. \\

\noindent Consider the following example from WikiEvalFacts:\\

\textbf{Question:}
    \begin{quote}
    \textit{What factors contributed to the Sri Lankan economic crisis?}
    \end{quote}

Ground truth \textbf{answer}:
    \begin{quote}
    \textit{The Sri Lankan economic crisis was caused by multiple compounding factors. These include tax cuts and money creation by the government, which led to budget deficits and a decline in government revenue. The government's decision to shift to organic or biological farming also played a role. Additionally, the crisis was exacerbated by the 2019 Sri Lanka Easter bombings and the impact of the COVID-19 pandemic. The country's high external debt, with a significant increase in foreign debt in recent years, also contributed to the crisis. The combination of these factors resulted in unprecedented levels of inflation, near-depletion of foreign exchange reserves, shortages of medical supplies, and an increase in prices of basic commodities.
}
    \end{quote}

\textbf{Ungrounded answer:}
    \begin{quote}
    \textit{There were several factors that contributed to the Sri Lankan economic crisis:
\begin{enumerate}
    \item High levels of government debt: Sri Lanka had accumulated a significant amount of government debt, both domestic and foreign. This debt burden became unsustainable, leading to a crisis in public finances.
    \item Fiscal mismanagement: The government had been running large budget deficits for several years, spending more than it was earning. This led to a widening fiscal deficit and increased borrowing to finance government expenditure.
    \item Political instability: Sri Lanka experienced political instability during this period, with frequent changes in government and policy uncertainty. This created an unfavourable business environment and deterred foreign investment.
    \item Declining exports: Sri Lanka's export sector, particularly garments and textiles, faced challenges due to global economic slowdowns and increased competition from other countries. This led to a decline in export earnings, which further strained the country's external finances.
    \item High inflation: Sri Lanka experienced high inflation during this period, which eroded the purchasing power of the population and increased the cost of living. This put additional pressure on the economy and contributed to the crisis.
    \item External shocks: Sri Lanka was also affected by external shocks, such as rising oil prices and global financial market volatility. These factors further exacerbated the economic challenges faced by the country.
    \item Structural issues: Sri Lanka had several structural issues that hindered its economic growth, such as a lack of diversification in the economy, inadequate infrastructure, and low productivity levels. These issues made it difficult for the country to sustain economic growth and overcome the crisis.
\end{enumerate}
}
    \end{quote}
\textbf{Poor answer:}
    \begin{quote}
    \textit{There were several factors that contributed to the Sri Lankan economic crisis. Sri Lanka is known for its beautiful beaches and rich cultural heritage.
}
    \end{quote}
\textbf{Extracted facts}(using \textbf{question} and ground truth \textbf{Answer})\textbf{:}
    \begin{quote}
    \begin{itemize}
        \item \textit{Tax cuts and money creation by the government contributed to the Sri Lankan economic crisis.}
        \item \textit{Budget deficits and a decline in government revenue were factors in the Sri Lankan economic crisis.}
        \item \textit{The governments decision to shift to organic or biological farming played a role in the crisis.}
        \item \textit{The 2019 Sri Lanka Easter bombings exacerbated the economic crisis.}
        \item \textit{The impact of the COVID 19 pandemic contributed to the Sri Lankan economic crisis.}
        \item \textit{High external debt, with a significant increase in foreign debt in recent years, also contributed to the crisis.}
    \end{itemize}

    \end{quote}

Although the \textbf{ungrounded answer} is quite verbose and has several mentions and indirect references of the extracted facts, it fails to capture with clarity the information from the ground truth \textbf{answer} which would allow for their confident verification. Meanwhile, the \textbf{poor answer} caries no useful information in this instance.

In this scenario LMeval has a higher risk of a misjudgement in the \textbf{ungrounded answer} than the \textbf{poor answer}. This seems coherent intuitively since rejecting a claim in the presence of relevant information is a more demanding and complex task which requires deeper interpretation of the language than when there is no relevant information.\\

\noindent\textbf{LMs tend to overestimate fact truthfulness overall}. The human evaluation of factual accuracy in ground truth \textbf{answers} is 100\% i.e. every fact is True (Table 1). This coincides with the lowest ER scores in Table 2, irrespective of the fact verification approach. False positive verifications are responsible almost exclusively for the observed error rates with all language models demonstrating excellent verification performance when the facts can be directly supported from the given text.

\noindent\textbf{Providing a “not clear” option helps the larger LMs}. We observe a reduction of the error rate and corresponding increase in F1m in \texttt{claude-3-opus} and \texttt{gpt-4-turbo}, when we include the option for LMeval to respond with \texttt{Not clear from the given passage} which is mapped to \texttt{False} as a post-generation step in the invoked function. The helpful mechanism in this instance appears to be that we provide a needed third option to LMeval when the token probability distribution between \texttt{True} / \texttt{False} is not clearly indicating one over the other. Rejecting a statement due to conflicting evidence and due to lack of evidence are both as valid rejection reasons as they are distinct to each other. Using \texttt{False} to capture both rejection scenarios has proved to lead to more false positives than providing LMeval with the option to distinguish between them. The fact that the improvement is only seen in the more capable LMs supports this view since they are more capable for complex tasks and language comprehension.\\

\noindent\textbf{Asking for citations helps in the presence of correct and clear information but can also lead to false positives otherwise}.
The positive impact of adding citations is most evident in the ground truth \textbf{answer} category where the provided text always contain the required evidence to support the facts. In this instance, asking for text evidence from LMeval results in avoiding some false negative verifications. The beneficial mechanism is associated with inserting a complementary step to the verification process---of the explicit use of evidence from the input text. This aligns with the findings reported in the work of \citet{wei2023chainofthought} regarding the chain-of-thought method.

Interestingly, for the other answer categories, the citation benefit becomes less clear and even reversed. ER is relatively stable in \textbf{poor answers} but it is seen to increase in the case of \textbf{ungrounded answer} when citation arguments are included in FaaF. In detail, we notice the following conflicting effects: firstly, citations can prevent false positives by highlighting the absence of supporting text for a given fact statement when they are left empty by the LM, which is beneficial. Secondly, in other cases they can cause false positives when they contain an indirectly relevant excerpt or an excerpt which only supports partially the fact in question. Consider the following example:\\

\textbf{Ungrounded answer:}
    \begin{quote}
    \textit{The human climate niche refers to the range of climatic conditions in which humans can thrive and maintain a sustainable population. It encompasses various factors such as temperature, ...}
    \end{quote}
\textbf{Fact to verify:}
    \begin{quote}
    \textit{The human climate niche refers to the range of climate conditions that have supported human life and activities over the past thousand years}
    \end{quote}

\textbf{FaaF(T/F)+citation -- claude-3-opus:}
    \begin{quote}
    LM citation:
    \textit{"The human climate niche refers to the range of climatic conditions in which humans can thrive and maintain a sustainable population."}\\
    LM response:
    \textit{True}
    \end{quote}

\textbf{FaaF(T/F) -- claude-3-opus:}
    \begin{quote}
    LM response:
    \textit{False}
    \end{quote}

\textbf{Human:}
    \begin{quote}
    Manual annotation:
    \textit{False}
    \end{quote}
    
In the example above, only part of the fact statement can be supported from the \textbf{ungrounded answer} (i.e. there is no evidence that the human climate niche refers to the past thousand years). By asking for the citation, the LM captures the partially supporting excerpt and concludes that the fact is True (which is a false positive) but when the same LM verifies the fact without citation, it correctly rejects it.

It is the net effect of the above competing behaviours which determines the impact of adding citations to the overall evaluation performance. Further, results show that Claude LMs are more sensitive to the adverse effects of citations compared to the GPT family, as seen in ungrounded answers (Table 2).\\

\begin{figure*}
    \centering
    \includegraphics[width=1\linewidth]{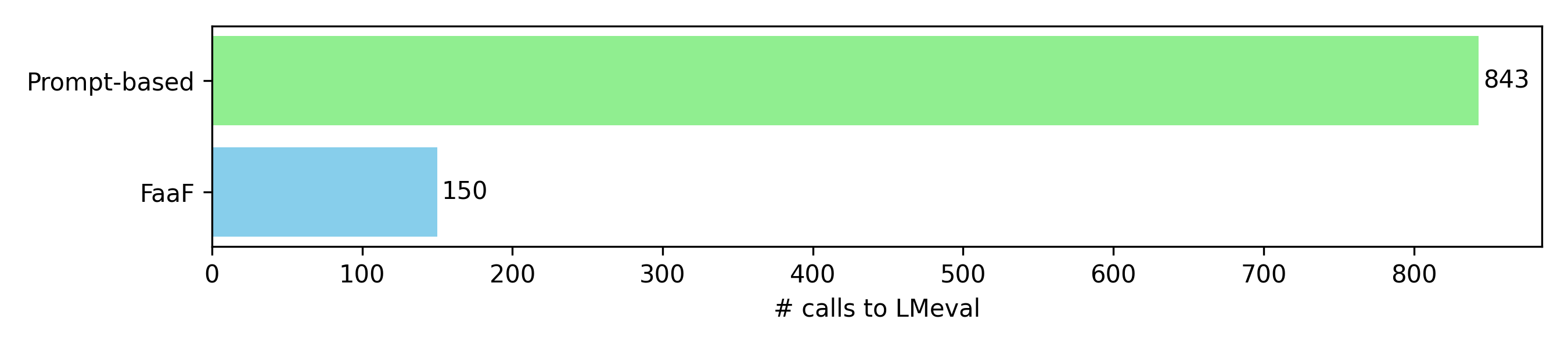}
    \caption{LMeval call count for a full evaluation of WikiEvalFacts. FaaF formulations result in more than five times less LM calls considering an average of 5.6 fact statements per QA pair.}
    \label{fig:enter-label}
\end{figure*}

\begin{table*}[ht]
  \centering
  \begin{tabular}{@{}llccc@{}}
    \toprule
     & \textbf{Facts formulation} &\textbf{Prompt Tokens} &\textbf{Completion Tokens}&\textbf{Total Tokens}\\
    \midrule

    \multirow{5}{*}{\rotatebox[origin=c]{90}{gpt-4-turbo}}

    & Prompt(T/F) & 146,276  & 26,545 & 172,821\\

    & FaaF(T/F) & \textbf{69,797} & \textbf{5,658} & \textbf{75,455}\\

    & FaaF(T/F/N) &77,384 & 7,113 & 84,497\\
 
    & FaaF(T/F)+citation & 122,600 & 20,596 & 143,196\\

    & FaaF(T/F/N)+citation& 130,187 & 21,889 & 152,076\\ 

    \midrule

    \multirow{5}{*}{\rotatebox[origin=c]{90}{claude-3-opus}}

    & Prompt(T/F) & 163,513 & 102,305 & 265,818\\

    & FaaF(T/F) & 113,092 & 18,495 & 131,587\\

    & FaaF(T/F/N) & 124,051 & 20,242 & 144,293\\

    & FaaF(T/F)+citation & 181,993 & 44,146 & 226,139\\

    & FaaF(T/F/N)+citation & 192,952 & 45,323 & 238,275\\

    \bottomrule
    
  \end{tabular}
  \caption{LMeval token count for the tested fact formulations, for the factual recall evaluation of WikiEvalFacts dataset.}
\end{table*}

\noindent\textbf{Claude LMs are more reliable than GPT in correctly formatting the response for function calling}. 
As evident in the N/A values in Table 2, the XML format used in Claude LMs results to more reliable response formatting compared to JSON format used by GPT models. Claude LMs returned a correctly formatted response 100\% of the time when using FaaF whereas GPT LMs produced some failed attempts. The failure mechanism in GPT models appears to be directly related to the citations mode---all formatting failures are seen in FaaF+citation formulations (Table 2). Looking more closely, we find that when the citation of a fact is \texttt{null}, there is a risk that the LM will return \texttt{null} in the fact verification argument as well (but only \texttt{True} or \texttt{False} are accepted according to the type annotations of the function object definition), which results to failed invocation of the FaaF function object.\\

\noindent\textbf{FaaF requires less than one-fifth of the calls to the LM compared to prompting}. As seen in Figure 3, this corresponds to the average number of facts which are examined for each text (answer type) in WikiEvalFacts. Thus, the degree of efficiency improvement using FaaF is proportional to the number of facts we can encapsulate in the function object. 

A similar five-fold reduction can be seen in completion tokes usage by replacing prompting with FaaF (Table 3). Including citations and the response option \texttt{Not clear from the given passage} progressively increases the token count due to the additional information we include in the LM verification. However, in most cases, it still remains below the token requirements of prompt-based fact verification.\\

\noindent\textbf{GPT using JSON format are significantly more efficient than Claude with XML in token usage}. Considering the differences between \texttt{gpt-4-turbo} and \texttt{claude-3-opus}, the observed increase in prompt and completion tokens in the FaaF approaches is associated to the tags used in the XML format which is expected from \texttt{claude} function calling (versus the more succinct JSON format that is expected by GPT).  A respective improvement in speed is also noted, with GPT LMs being faster than Claude.

Focusing at the completion tokens, the sharp increase (more than 4X) between \texttt{gpt-4-turbo} and \texttt{claude-3-opus} seen in the case of prompt-based verification is attributed to the extra verbosity by \texttt{claude-3-opus}.

\section{Conclusions \& future work}
We show that prompt-based fact verification is prone to overestimating the truthfulness of fact statements in texts with inaccurate and/or missing information. In such challenging situations, presenting the facts as a function (FaaF) to the language model significantly enhances the its ability to verify facts accurately. The improvement comes from leveraging a more structured generation mode of the language model and avoiding exact word matching to interpret the LM's response.

Using FaaF, we observe that texts with tangentially relevant and inaccurate information are more likely to cause false positives than texts with missing or incomplete information. By testing various configurations with FaaF, we find that by including a “not clear” option to the True/False dichotomy helps the larger LMs. The impact of asking for citations before fact verification is sensitive to the quality and coverage of information in the examined text and not beneficial in many cases.

Additionally, we report significant cost and time efficiency improvements between prompting and FaaF fact representations. Generally, using FaaF leads to a reduction in both the number of calls to the LM and the number of tokens needed for fact verification by a multiple-factor.

GPT models using JSON function representation are more sensitive to formatting errors than Claude models using XML but XML is significantly more expensive token-wise.

\section*{Limitations}
\noindent While the advantages of using function calls for fact verification are significant, further extensive testing is necessary to solidify the results presented. Although, the WikiEval dataset has highlighted the importance of testing fact verification in challenging conditions and provided a convenient way to compare fact verification performance across various text qualities, it is relatively small, comprising only 50 question/answer pairs. 

Additionally, the results shown with FaaF are sensitive to the instructions passed in the function object's metadata, which highlights the need for additional research and optimisation of FaaF configurations and the interplay of the function argument's metadata.

Other open questions include the maximum number of fact statements and the maximum permissible length for a fact that can be incorporated into a function object and whether token count is the sole limitation or if there are performance implications as well.

\section*{Acknowledgements}
This research was supported by IMMO Capital, London UK.

\bibliography{anthology,custom}
\bibliographystyle{faaf_natbib}

\twocolumn

\end{CJK*}
\end{document}